\documentclass{article}
\pdfoutput=1
\usepackage{times}
\usepackage{graphicx} 
\usepackage{subfigure}
\usepackage{natbib}
\usepackage{amssymb}
\usepackage{amsmath}
\usepackage{algorithm}
\usepackage{algorithmic}
\graphicspath{{figures/}}

\usepackage[accepted, nohyperref]{icml2015}

\begin{document}

\twocolumn[
\icmltitle{Rank Subspace Learning for Compact Hash Codes}

\icmlauthor{Kai Li}{kaili@eecs.ucf.edu}
\icmladdress{University of Central Florida,
            4000 Central Florida Blvd., Orlando, FL 32826 USA}
\icmlauthor{Guojun Qi}{guojun.qi@ucf.edu}
\icmladdress{University of Central Florida,
            4000 Central Florida Blvd., Orlando, FL 32826 USA}            
\icmlauthor{Jun Ye}{jye@eecs.ucf.edu}
\icmladdress{University of Central Florida,
            4000 Central Florida Blvd., Orlando, FL 32826 USA}
\icmlauthor{Kien A. Hua}{kienhua@eecs.ucf.edu}
\icmladdress{University of Central Florida,
            4000 Central Florida Blvd., Orlando, FL 32826 USA}

\vskip 0.3in
]

\begin{abstract}
The era of Big Data has spawned unprecedented interests in developing hashing algorithms for efficient storage and fast nearest neighbor search. 
Most existing work learn hash functions that are numeric quantizations of feature values in projected feature space. In this work, we propose a novel hash learning framework that encodes feature's rank orders instead of numeric values in a number of optimal low-dimensional ranking subspaces. We formulate the ranking subspace learning problem as the optimization of a piece-wise linear convex-concave function and present two versions of our algorithm: one with independent optimization of each hash bit and the other exploiting a sequential learning framework. Our work is a generalization of the Winner-Take-All (WTA) hash family and naturally enjoys all the numeric stability benefits of rank correlation measures while being optimized to achieve high precision at very short code length. We compare with several state-of-the-art hashing algorithms in both supervised and unsupervised domain, showing superior performance in a number of data sets.
\end{abstract}

\section{Introduction}
\label{intro}

Massive amount of social multimedia data are being generated by billions of users every day. The advent of multimedia big data presents a number of challenges and opportunities for research and development of efficient storage, indexing and retrieval techniques. Hashing is recognized by many researchers as a promising solution to the above Big Data problem, thus attracting significant amount of research in the past few years \cite{lsh, bre, rsh, sh}. Most hashing algorithms encode high-dimensional data into binary codes by quantizing numeric projections \cite{mlh, anchor, dgh, ksh}. In contrast, hashing schemes based on feature's ranking order (i.e. comparisons) are relatively underresearched and will be the focus of this paper.

Ranking-based hashing, such as Winner-Take-All (WTA) \cite{wta} and Min-wise Hashing (MinHash) \cite{minhash}, ranks the random permutation of input features and uses the index of maximal/minimal feature dimensions to encode a compact representation of the input features. The benefit of ranking-based hashing lies in the fact that these algorithms are insensitive to the magnitude of features, and thus are more robust against many types of random noises universal in real applications ranging from information retrieval \cite{rbm}, image classification \cite{jsd} to object recognition \cite{rbmimg}.  In addition, the magnitude-independence also makes the resultant hash codes scale-invariant, which is critical to compare and align the features from heterogeneous spaces, e.g., revealing the multi-modal correlations \cite{self}.

Unfortunately, the existing ranking-based hashing is data-agnostic. In other words, the obtained hash codes are not learned by exploring the intrinsic structure of data distribution, making it suboptimal in its efficiency of coding the input features with compact codes of minimal length.  For example, WTA encodes the data with the indices of the maximum dimensions chosen from a number of random permutations of input features.  Although WTA has generated leading performances in many tasks \cite{cvprbest,wta,self}, it is constrained in the sense that it only ranks the existing features of input data, while incapable of combining multiple features to generate new feature subspaces to rank. A direct consequence of such limitation is that this sort of ranking-based hashing usually needs a very large number of permutations and rankings to generate useful codes, especially with a high dimensional input feature space \cite{wta}.

To address this challenge, we abandon the use of ranking random permutations of existing features in ranking-based hashing algorithms. Instead, we propose to generate compact ranking-based hashing codes by learning a set of new subspaces and ranking the newly projected features in these subspaces.  At each step, an input data is encoded by the index of the maximal value over the projected points onto these subspaces. The subspace projections are jointly optimized to generate the ranking indices that are most discriminative to the metric structure and/or the data labels. Then a vector of codes are iteratively generated to represent the input data from the maximal indices over a sequence of sets of subspaces.

This method generalizes ranking-based hashing from restricted random permutations to perform encoding by ranking a set of arbitrary subspaces learned by mixing multiple original features.  This greatly extends its flexibility so that much shorter bits can be generated to encode input data, while retaining the benefits of noise insensitivity and scale invariance inherent in such algorithms.

In the remainder of this paper, we first review the related hashing algorithms in Section \ref{related}, then the rank subspace hash learning problem is formulated and solved in Section \ref{form}. In Section \ref{seq}, we present an improved learning algorithm based on sequential learning. The experimental results are presented in Section \ref{exp} and the paper is concluded in Section \ref{conclude}.

\section{Related Work}\label{related}
Since the focus of this paper is on data-dependent hashing, we limit our review to two categories of this line of research -- unsupervised and supervised hashing. Interested reader can refer to \cite{survey13} and \cite{survey14} for a comprehensive review of this research area.

The most representative works on unsupervised hashing includes Spectral Hashing \cite{sh} and kernelized variant Locality Sensitive Hashing \cite{klsh}. In detail, SH learns linear projections through an eigenvalue decomposition so that distance between pairs of similar training samples are minimized in the projected subspaces when these samples are binarized. Similarly, KLSH also makes use of an eigensystem solution, but it manipulates data items in the kernel space in an effort to generalize LSH to accommodate arbitrary kernel functions. Binary Reconstructive Embedding (BRE) \cite{bre} explicitly minimizes the reconstruction error between the input space and Hamming space to preserve the metric structure in input space, which has demonstrated improved performance over SH and LSH. Iterative Quantization (ITQ) \cite{itq} iteratively learns uncorrelated hash bits to minimize quantization error between hash bodes and the dimension-reduced data.
\cite{splh} presents a sequential projection learning method that fits the eigenvector solution into a boosting framework and uses pseudo labels in learning each hash bit. More recently, Anchor Graph Hashing (AGH) \cite{anchor} and Discrete Graph Hashing (DGH) \cite{dgh} use anchor graphs to capture the neighborhood structure inherent in a given dataset and adopt a discrete optimization procedure to achieve nearly balanced and uncorrelated hash bits.

On the other hand, supervised hashing methods take advantage of the data labels to learn data-dependent hash functions. It has been shown that supervised hashing methods are good at incorporating data labels to learn more discriminative hash codes in many tasks. For example, \cite{rbm} uses Restricted Boltzman Machine (RBM) to learn nonlinear binary hash codes for document retrieval and demonstrates better precision and recall than other methods. Similar deep learning hash methods have also been applied to the task of image retrieval in very large databases \cite{rbmimg}. However, deep learning methods typically need large data sets, cost long training times and have been outperformed by other methods exclusively designed for learning hashing functions. For instance, \cite{mlh} proposes the Minimal Loss Hashing (MLH) with a structural SVM-like formulation and minimizes the loss-adjusted upper bound of a hinge-like loss function defined on pairwise similarity labels. The resulting hash codes have shown to give superior performance over the state-of-the-art. This method is further extend to minimize loss functions defined with triplet similarity comparisons \cite{hamdist}. Similarly, \cite{cghash} also learns hash functions based on triplet similarity. On the contrary, the formulation is a convex optimization within the large-margin learning framework rather than structural SVM.

Recently, \cite{jsd} theoretically proved the convergence properties of arbitrary sequential learning algorithms and proposed the Jensen Shannon Divergence (JSD) sequential learning method with a multi-class classification formulation. Supervised Hashing with Kernels (KSH) is another sequential learning algorithm.  This method maps the data to compact hash codes by minimizing Hamming distances of similar pairs and maximizing that of dissimilar pairs simultaneously \cite{ksh}.  The sequential part of our algorithm is similar to Boosting Similarity Sensitive Coding (BSSC) \cite{bssc} and Forgiving Hash (FH) \cite{fghash}, both of which treat each hash bit as a week classifier and learn a series of hash functions in a AdaBoost framework. However, the rank-based hash function we learn at each step is significantly different from that in BSSC and FH, resulting in completely different objective functions and learning steps. Existing hashing schemes based on rank orders (e.g. \cite{rsh}, \cite{pm} and \cite{keypoint}) are mostly restricted to approximating nearest neighbors given a distance metric to speed up large scale lookup. To the best of our knowledge, there has been no previous work explicitly exploiting the rank-based hash functions in a supervised hash learning setting.

\section{Formulation}
\label{form}
\subsection{Winner-Take-All Hashing}
The WTA hashing is a subfamily of hashing functions introduced by \cite{wta}. WTA is specified by two parameters: the number of random permutations $L$ and the window size $K$. Each permutation $\boldsymbol\pi$ rearranges the entries of an input vector $\mathbf x\in\mathbb R^d$ to $\mathbf x_{\pi}$ in the order specified by $\boldsymbol\pi$. Then the index of the maximum dimension of the feature among the first $K$ elements of $\mathbf x_{\pi}$ is used as the hash code.  This process is repeated $L$ times, resulting in a $K$-nary hash code of length $L$, which can be compactly represented using $L\times \lceil\log_2K\rceil$ bits.

WTA is considered as a ranking-based hashing algorithm, which uses the rank order among permuted entries of a vector rather than their values of features.  This property has given WTA certain degree of stability to perturbations in numeric values. Thus the WTA hash codes usually generate more robust metric structure to measure the similarity between input vectors than other types of hash codes which often contain inherent noises from quantizing the input feature spaces.  With theoretical soundness, however, the hash codes generated by WTA often must be sufficiently long to represent the original data in high fidelity.

This is caused by twofold limitations: (1) the entries of input vectors are permuted in a random fashion before the comparison is applied to find the largest entry out of the first $K$ ones; (2) the comparison and the ranking are restricted to be made between the original features. The random permutations are very inefficient to find the most discriminative entries to compare the similarity between the input vectors, and the restriction of the ranking to original features is too strong to generate the compact representations.  In the next, we relax the two limitations.

\subsection{Rank Subspace Hashing}

Rather than randomly permuting the input data vector $\mathbf{x}$, we project it onto a set of $K$ one-dimensional subspaces.  Then the input vector is encoded by the index of the subspace that generates the largest projected value. In other words, we have
\begin{equation} \label{eq1}
h(\mathbf{x};\mathbf{W}) = \arg\max_{1\leq k\leq K} \mathbf{w}_k^T\mathbf{x},
\end{equation}
where $\mathbf{w}_k\in\mathbf R^d, 1\leq k\leq K$ are vectors specifying the subspace projections, and $\mathbf{W} = [\mathbf{w}_1, \mathbf{w}_2, \cdots, \mathbf{w}_K]^T$.

We use a linear projection to map an input vector into subspaces to form its hash code.  At first glance, this idea is similar to the family of learning-based hashing algorithms based on linear projection \cite{lsh,mlh}.  However, different from these existing algorithms, the proposed method instead ranks the obtained subspaces to encode each input vector with the index of the dimension with the maximum value. This makes the obtained hash codes highly nonlinear to the input vector, invariant to the scaling of the vector, as well as insensitive to the input noises to a larger degree than the linear hashing codes.  In this paper, we name this method Rank Subspace Hashing (RSH) to distinguish it from the other compared methods.

WTA is a special case of the RSH algorithm, if we restrict the projections onto $K$ axis-aligned linear subspaces, i.e., $\mathbf{w}_k$ is set to a column vector $\mathbf e_k$ randomly chosen from an identity matrix $\mathbf I$ of size $d\times d$.

RSH extends WTA by relaxing the axis aligned linear subspaces in (\ref{eq1}) to arbitrary $K$-dimensional linear subspaces in $\mathbb{R}^d$. Such relaxation greatly increases the flexibility to learn
a set of subspaces to optimize the hash codes resulting from the projections to these subspaces.

Now our objective boils down to learn hash functions characterized by the projections $\mathbf W$ as in Eq. (\ref{eq1}).  Specifically, let $\mathcal{D}$ be the set of $N$ $d$-dimensional data points $\{\mathbf{x}_i\}_{i=1}^N$ and let $\mathcal{S}=\{s_{ij}\}_{1\leq i,j\leq N}$ be the set of pair-wise similarity labels satisfying $s_{ij}\in \{0, 1\}$, where $s_{ij} = 1$ means the pair $(\mathbf{x}_i, \mathbf{x}_j)$ is similar and vice verse. The pair-wise similarity labels $\mathcal{S}$ can be obtained either from the nearest neighbors in a metric space or by human annotation that denotes whether a pair of data points come from the same class.

Given a similarity label $s_{ij}$ for each training pair, we can define an error incurred by a hash function like (\ref{eq1}) below
\begin{align}\label{err}
e(h_i, h_j, s_{ij}) = \left\{
\begin{array}{lr}
\rho I(h_i \neq h_j) , & s_{ij} = 1\\
\lambda (1 - I(h_i \neq h_j)) , & s_{ij} = 0
\end{array} \right.
\end{align}
where $I(\cdot)$ is the indicator function outputting 1 when the condition holds and 0 otherwise, $h_{i(j)}$ is $h(\mathbf{x}_{i(j)}; \mathbf{W})$ for short, and $\rho$ and $\lambda$ are two hyper-parameters that penalize false negative and false positive respectively.

The learning objective is to find $\mathbf W$ to minimize the cumulative error function over the training set:
\begin{equation}\label{obj}
E(\mathbf{W}) = \sum_{1\leq i \leq j \leq N} e({h}_i, {h}_j, s_{ij})
\end{equation}
Note that $\mathbf W$ factors into the above objective function because both ${h}_i$ and ${h}_j$ are a function of $\mathbf W$.

\subsection{Reformulation}

The above objective function is straightforward to formulate, but hard to optimize because it involves the indicator function and $\arg\max$ function which are typically non-convex and highly discontinuous. Motivated by \cite{mlh}, we reformulate the objective function and seek a piecewise linear upper bound of $E(\mathbf{W})$.

First, the hash function in (\ref{eq1}) can be equivalently reformulated as
\begin{equation}\label{scorefunc}
\begin{aligned}
&\mathbf h(\mathbf{x};\mathbf{W}) = \arg\max_\mathbf{g} \mathbf{g}^T\mathbf{Wx},\\
&{\rm subject~~to~~} \mathbf g \in \{0,1\}^K, \mathbf 1^T \mathbf g = 1,
\end{aligned}
\end{equation}
which outputs an 1-of-$K$ binary code $\mathbf h$ for an input feature vector $\mathbf x$.  The constraint enforces there must exist and only exist
a nonzero entry of $1$ in the resultant hash code. We enforce this constraint in the following optimization problems without meaning it explicitly to avoid notational clutter.
It is easy to find the equivalence to the hashing function (\ref{eq1}): the only nonzero element in $\mathbf h$ encodes the index of dimension with the maximum value in $\mathbf{Wx}$.

Given a pairwise similarity label $s_{ij}$ between two vectors $\mathbf x_i$ and $\mathbf x_j$, $\mathbf{h}_i$ and $\mathbf{h}_j$ are their hash codes obtained by solving the $\arg\max$ problem (i.e. $\mathbf{h}(\mathbf x_i;\mathbf W)$ and $\mathbf{h}(\mathbf x_j;\mathbf W)$). Then the error function (\ref{eq1}) can be upper bounded by
\begin{equation}\label{upbound}
\begin{aligned}
e(\mathbf{h}_i,\mathbf{h}_j,s_{ij}) \leq &\max_{\mathbf{g}_i, \mathbf{g}_j}[e(\mathbf{g}_i,\mathbf{g}_j,s_{ij})+\mathbf{g}_i^T\mathbf{Wx}_i+\mathbf{g}_j^T\mathbf{Wx}_j] \nonumber\\
&-\mathbf{h}_i^T\mathbf{Wx}_i-\mathbf{h}_j^T\mathbf{Wx}_j
\end{aligned}
\end{equation}
This inequality is easy to prove by noting that the following inequality
\[
\begin{aligned}
\max&_{\mathbf{g}_i, \mathbf{g}_j}[e(\mathbf{g}_i,\mathbf{g}_j,s_{ij})+\mathbf{g}_i^T\mathbf{Wx}_i+\mathbf{g}_j^T\mathbf{Wx}_j] \\
&\geq e(\mathbf{h}_i,\mathbf{h}_j,s_{ij})+\mathbf{h}_i^T\mathbf{Wx}_i+\mathbf{h}_j^T\mathbf{Wx}_j
\end{aligned}
\]

With the above upper bound of error function, we seek to solve the MinMax problem of minimizing the following function with respect to $\mathbf{W}$
\begin{equation}\label{obj}
\begin{aligned}
\Omega(\mathbf W)=&\sum_{1\leq i<j\leq N}\{\max_{\mathbf{g}_i, \mathbf{g}_j}[e(\mathbf{g}_i,\mathbf{g}_j,s_{ij})+\mathbf{g}_i^T\mathbf{Wx}_i+\mathbf{g}_j^T\mathbf{Wx}_j] \nonumber\\
&-\mathbf{h}_i^T\mathbf{Wx}_i-\mathbf{h}_j^T\mathbf{Wx}_j\}
\end{aligned}
\end{equation}

\subsection{Optimization}
Consider $\mathbf{W}$ is fixed. The first step is a discrete optimization problem that is guaranteed to have global optimal solution. Specifically, given the values of $\mathbf{Wx}_{i(j)}$, the RSH codes $\mathbf h_{i(j)}$ in the second and third term of (\ref{obj}) can be found straightforwardly in $O(K)$.  For the adjusted error $e(\mathbf g_i, \mathbf{g}_j, s_{ij})+\mathbf{g}_i^T\mathbf{Wx}_i+\mathbf{g}_j^T\mathbf{Wx}_j$ of the first term in the square bracket, it is not hard to derive
its maximum value can be obtained by scanning the elements in matrix $[m_{kl}]_{K\times K}$, defined as
\begin{align}\label{adjerr}
m_{kl} = \left\{
\begin{array}{ll}
 y_i^{(k)}+y_j^{(l)} + \lambda (1-s_{ij}) & \mbox{if }k = l \\
 y_i^{(k)}+y_j^{(l)} + \rho s_{ij} & \mbox{otherwise}
\end{array} \right.
\end{align}
where $y_i^{(k)}$ is the $k^{th}$ element of $\mathbf{Wx}_i$. Assuming
the $(k^*,l^*)$th element of the above matrix achieves the maximum value, the maxima $(\mathbf g_i^*, \mathbf g_j^*)$ of the adjusted error are 1-of-$K$ binary vectors  with the $k^*$th and the $l^*$th dimension set to 1.
The above procedure can be computed in $O(K^2)$. Since $K$ is normally very small (e.g. 2 to 8), the above discrete optimization problem can be computed efficiently.

Now consider the optimization of $\mathbf{W}$.
Fixing the maxima $(\mathbf{g}_i^*, \mathbf{g}_j^*)$ of the first term, and the RSH codes ${ \mathbf{h}_i}$ and ${\mathbf{h}_j}$ in (\ref{scorefunc}), $\mathbf W$ can be updated in the direction of the negative gradient
\begin{equation}\label{grad}
-\frac{\partial \Omega(\mathbf{W})}{\mathbf{W}} = \sum_{i, j} ({\mathbf{h}_i}-\mathbf{g}_i^*)\mathbf{x}_i^T+({\mathbf{h}_j}-\mathbf{g}_j^*)\mathbf{x}_j^T
\end{equation}
Batch update can be made using (\ref{grad}) when the training data can be loaded into the memory all at once. Otherwise, $\mathbf{W}$ can also be done in an online fashion with one training pair at a time, leading to the following iterative learning procedure
\begin{equation}\label{update}
\mathbf W \leftarrow \mathbf W + \eta \big[ ({\mathbf{h}_i}-\mathbf{g}_i^*)\mathbf{x}_i^T+({\mathbf{h}_j}-\mathbf{g}_j^*)\mathbf{x}_j^T \big],
\end{equation}
where $\eta$ is the learning rate.

The learning algorithm is shown as Algorithm \ref{rsh}. The algorithm learns $L$ projection matrices by starting with different random initializations from Gaussian distribution. Because the convex-concavity nature of the objective function, the solutions have multiple local minima.  This is a desired property  in our application, because each local minimum, corresponding to a RSH function, reflects a distinct perspective of ranked subspaces underlying the training examples.
In addition, each hash function is learned independently and thus can be done in parallel.
The convergence of the learning algorithm has been explored and empirically studied in \cite{directloss, mlh}.

\begin{algorithm}[tb]
   \caption{Rank Subspace Learning}
   \label{rsh}
\begin{algorithmic}
   \STATE {\bfseries Input:} data $[\mathbf{x}_i]$, pairwise similarity labels $[s_{ij}]$, length of hash code $L$, subspace dimension $K$
	\FOR{$l=1$ {\bfseries to} $L$}
	\STATE Initialize $\mathbf{w}_k, 1\leq k\leq K$ from Gaussian distribution   	
	\REPEAT
	\STATE Pick a pair $(\mathbf{x}_i, \mathbf{x}_j)$ and compute ${\mathbf{h}}_i$, ${\mathbf{h}}_j$, $\mathbf{g}_i^*$, $\mathbf{g}_j^*$
	\STATE Update projection matrix $\mathbf{W}$ according to
	\STATE \center{$\mathbf{W} \leftarrow \mathbf{W} + \eta [({\mathbf{h}_i}-\mathbf{g}_i^*)\mathbf{x}_i^T+({\mathbf{h}_j}-\mathbf{g}_j^*)\mathbf{x}_j^T\big]$}
	\UNTIL Convergence
   	\ENDFOR
\end{algorithmic}
\end{algorithm}

\section{The Sequential Learning}
\label{seq}
In Algorithm \ref{rsh}, since each hash function is learned independently, the entire hash code may be suboptimal. This is because different random starting points may lead to the same local minima, resulting in redundant hash bits. In order to maximize the information contained in a $L$-bit hash code, we propose to learn the hash functions sequentially so each hash function can provide complementary information to previous ones.

In order to motivate our sequential learning algorithm, we can view each hash bit as a week classifier that assigns similarity labels to an input pair, and the obtained ensemble classifier is related with the Hamming distance between hashing codes. Formally, each week classifier corresponding to the $l^{th}$ bit is
\begin{equation}\label{sim}
sim_l(\mathbf{x}_i, \mathbf{x}_j) = 1-H_m(\mathbf h(\mathbf{x}_i;\mathbf W_l),\mathbf h(\mathbf{x}_j;\mathbf W_l))
\end{equation}
Where $H_m(x,y) = I(x\neq y)$ is the bitwise Hamming distance, and $\mathbf W_l$ is the projection matrix for this bit. Then, the  Hamming distance between two $L$-bit hash codes can be seen as the vote of an ensemble of $L$ week classifiers on them. Clearly, the sequential learning problem naturally fits into the AdaBoost framework.

The AdaBoost-based sequential learning algorithm is shown in Algorithm \ref{srsh}. In detail, a sampling weight $\alpha_{ij}^{(l)}$ is assigned to each training pair and is updated before training each new hash function. In particular, pairs that are misclassified by the current hash function will be given more weight in training the next hash function. 
 The projection matrix is updated in the similar online fashion as in (\ref{update}) but weighted by the sampling weight.
 
 When all the hash functions have been trained, the voting results of the related week classifiers are fused with a weighted combination
\begin{equation}
sim(\mathbf{x}_i, \mathbf{x}_j) = \sum_{l=1}^{L}\theta_l(1-H_m(\mathbf h(\mathbf x_{i};\mathbf W_l),\mathbf h(\mathbf x_{j};\mathbf W_l)),
\end{equation}
where $\theta_l$ are the weighted training error of the $l^{th}$ hash function.

We name this AdaBoost-inspired sequential learning by Sequential RSH (SRSH), 
in contrast to the RSH algorithm with independently composed hash codes. 

\begin{algorithm}[tb]
   \caption{Sequential Rank Subspace Learning}
   \label{srsh}
\begin{algorithmic}
   \STATE {\bfseries Input:} data $[\mathbf{x}_i]$, pairwise similarity labels$[s_{ij}]$, length of hash code $L$, subspace dimension $K$
   \STATE {\bfseries Initialize:} set all the sampling weights $\{\alpha_{ij}\}$ to 1
	\FOR{$l=1$ {\bfseries to} $L$}
	\STATE Initialize $\mathbf{W}_l$ from Gaussian distribution   	
	\REPEAT 
	\STATE Pick a pair $(\mathbf{x}_i, \mathbf{x}_j)$ and compute ${\mathbf{h}}_i$, ${\mathbf{h}}_j$, $\mathbf{g}_i^*$, $\mathbf{g}_j^*$ based on the  current estimate of $\mathbf W_l$;
		\STATE Update projection matrix $\mathbf{W}_l$ according to
	\STATE $\mathbf{W}_l \leftarrow \mathbf{W}_l + \eta\alpha_{ij}^{(l)} \big[({\mathbf{h}_i}-\mathbf{g}_i^*)\mathbf{x}_i^T+({\mathbf{h}_j}-\mathbf{g}_j^*)\mathbf{x}_j^T\big]$;
\UNTIL Convergence
	\STATE Compute the weighted errors
	\STATE $$\epsilon_l = \frac{\sum_{i,j} \alpha_{ij}^{(l)} e(\mathbf{h}_i, \mathbf{h}_j, s_{ij})}{\sum_{i,j} \alpha_{ij}^{(l)}}$$
	\STATE Evaluate the quantity
	\STATE $$\theta_l = \ln \Big\{\frac{1-\epsilon_l}{\epsilon_l}\Big\}$$
	\STATE Update the pair weighting coefficients using
	\STATE $$\alpha_{ij}^{(l+1)} \propto \alpha_{ij}^{(l)}\exp\{\theta_le(\mathbf{h}_i, \mathbf{h}_j, s_{ij})\}$$
	\STATE Normalize the sampling weights such that $\sum_{i,j}{\alpha_{ij}^{(l+1)}}=\sum_{ij}\alpha_{ij}^{(l)}$.
   	\ENDFOR
\end{algorithmic}
\end{algorithm} 

\section{Experiments}
\label{exp}

\begin{figure*}[ht]
\vskip 0.2in
\centering
\subfigure[LabelMe]{\label{ap1}\includegraphics[width=0.3\textwidth]{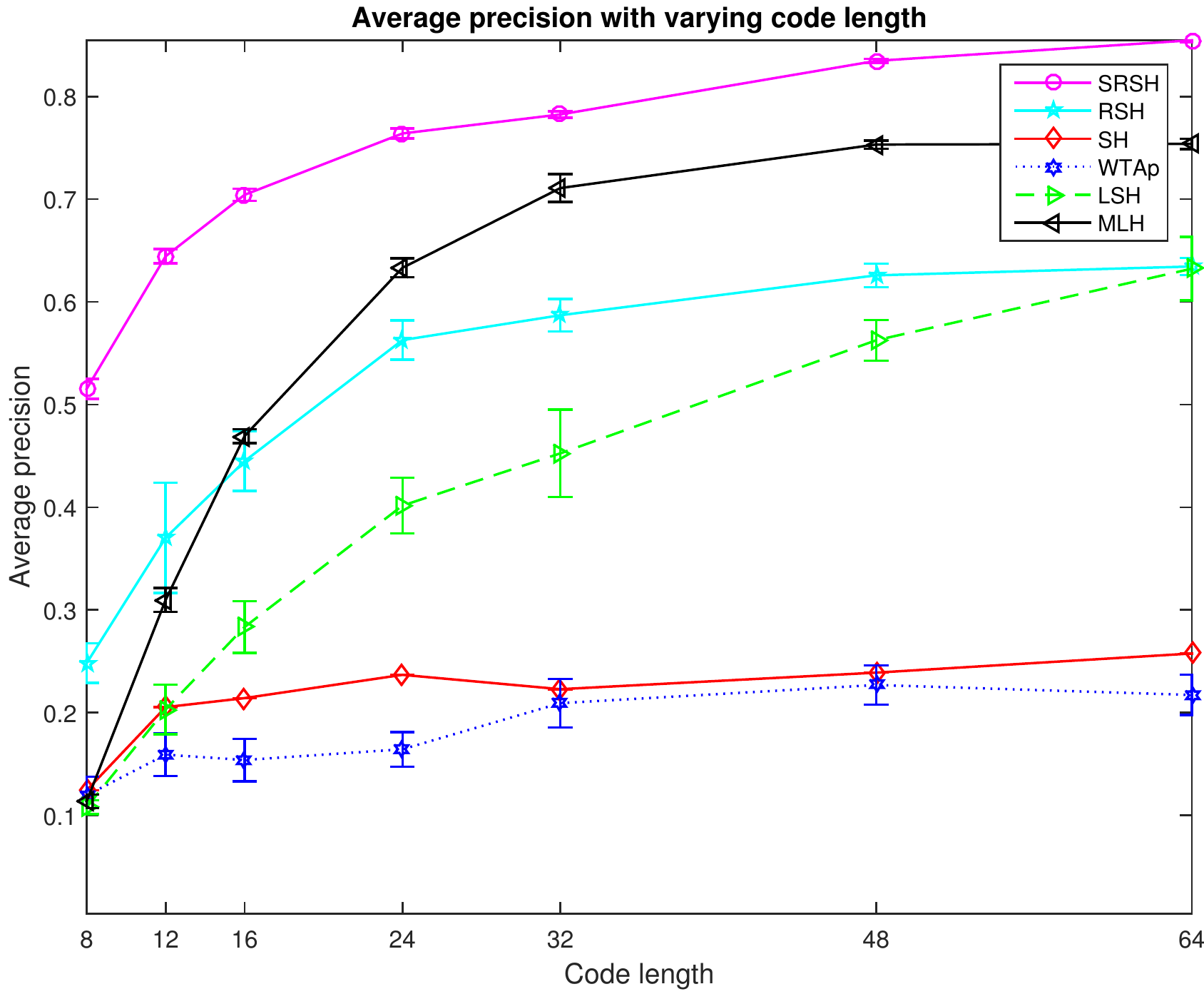}}
\subfigure[MNIST]{\label{ap2}\includegraphics[width=0.3\textwidth]{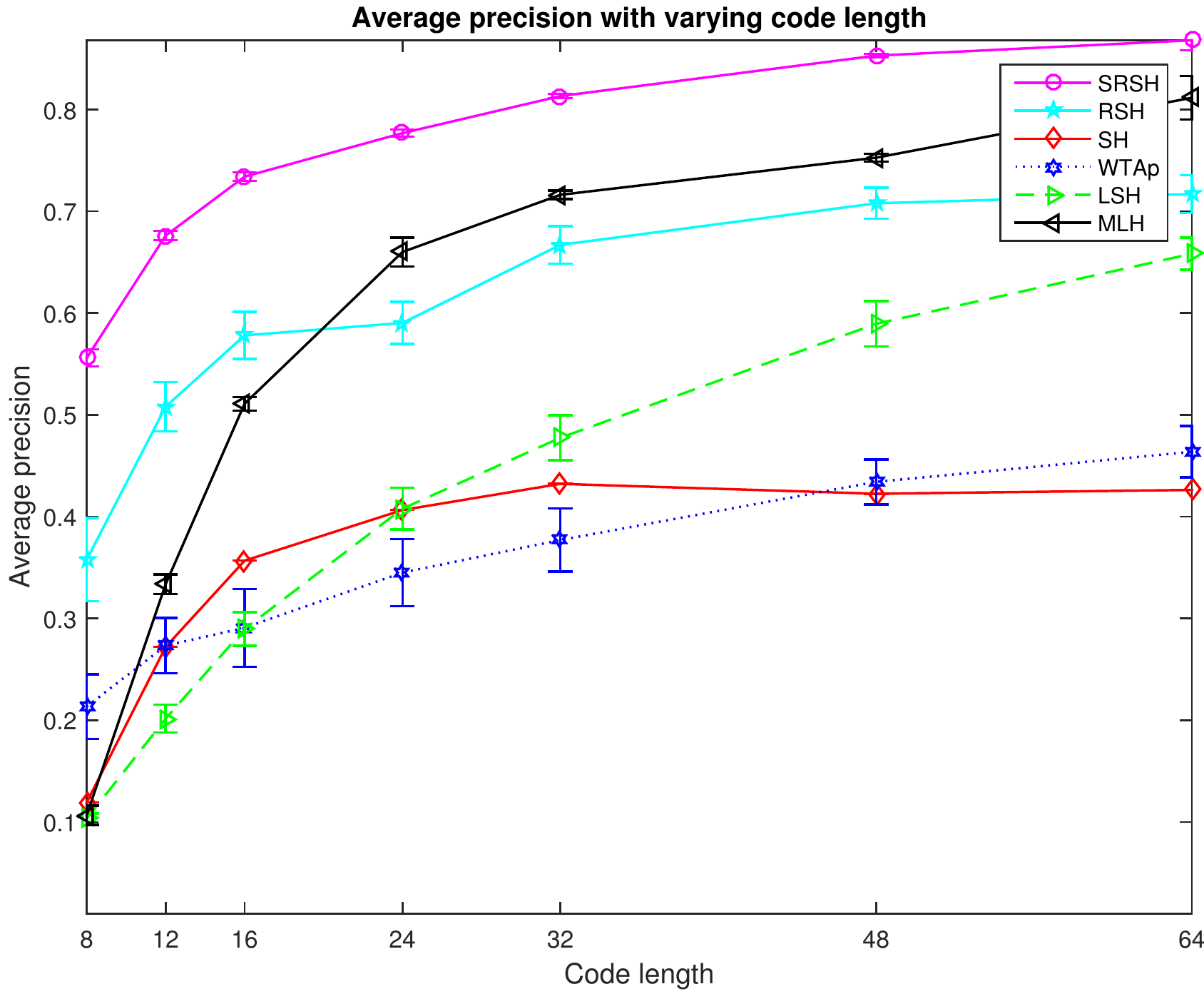}}
\subfigure[Peekaboom]{\label{ap4}\includegraphics[width=0.3\textwidth]{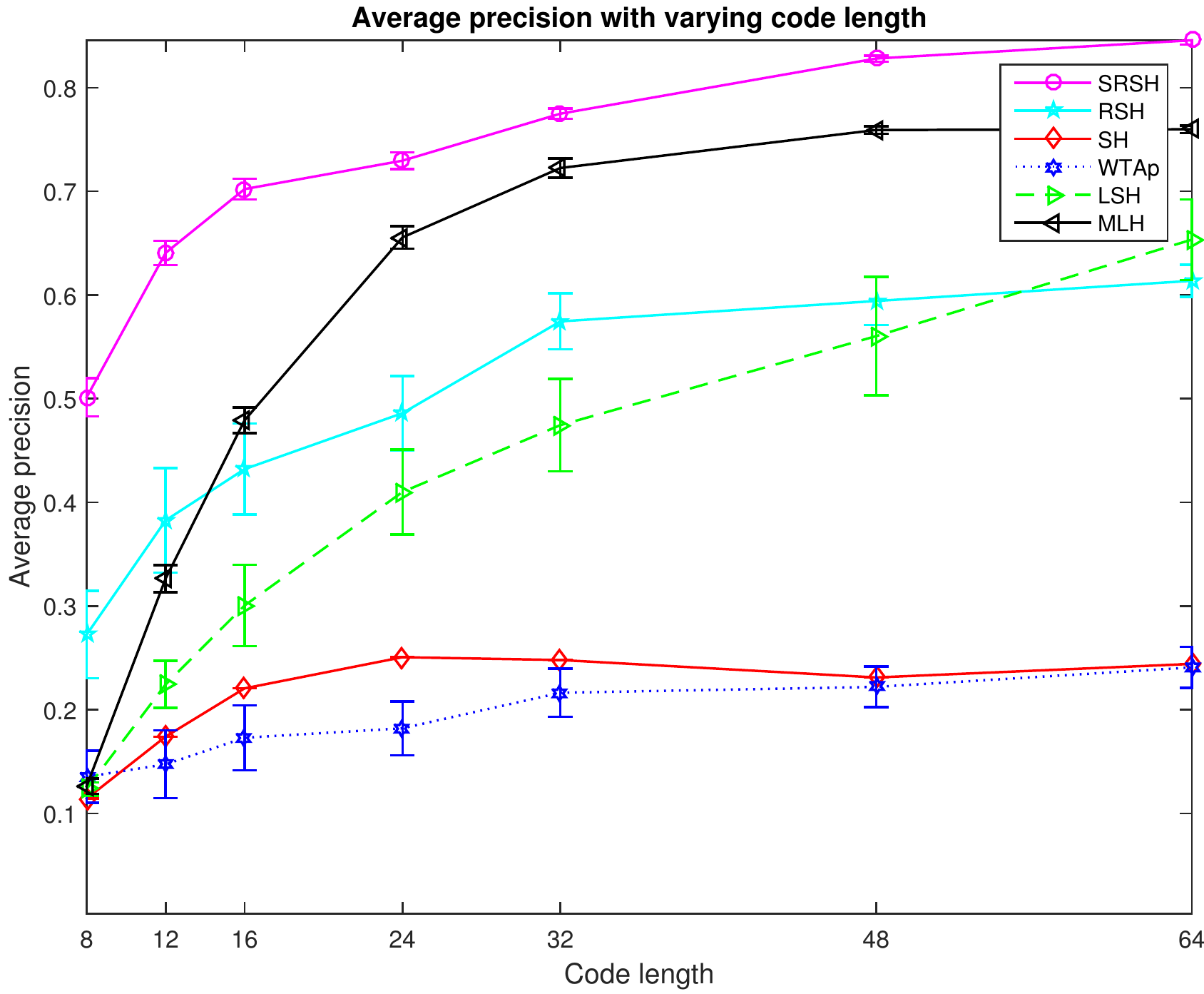}}
\caption{Average precision with varying hash code length.}
\label{ap}
\vskip -0.2in
\end{figure*}

\begin{figure*}[ht]
\vskip 0.2in
\centering
\subfigure[LabelMe]{\label{prrc1}\includegraphics[width=0.3\textwidth]{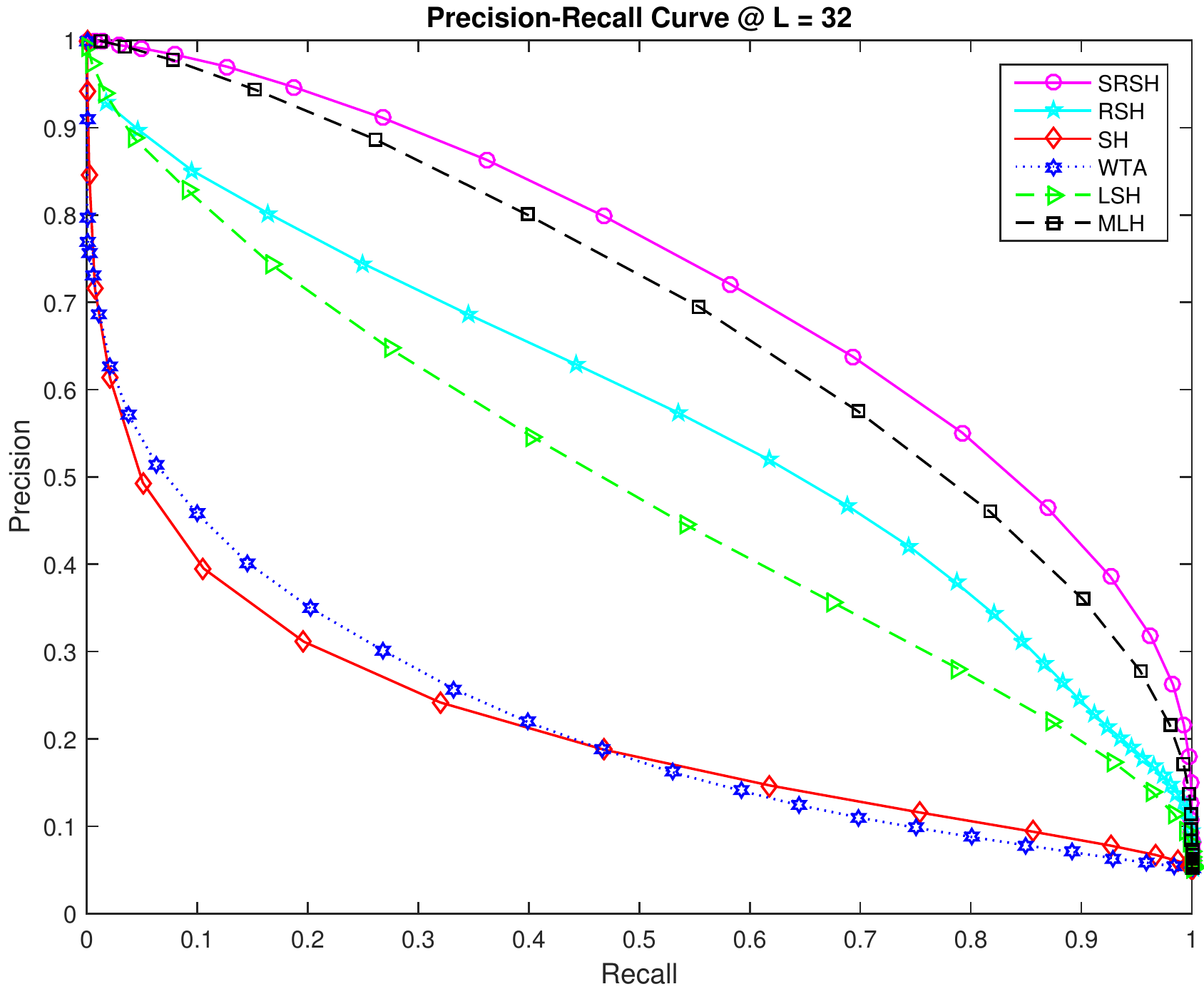}}
\subfigure[MNIST]{\label{prrc2}\includegraphics[width=0.3\textwidth]{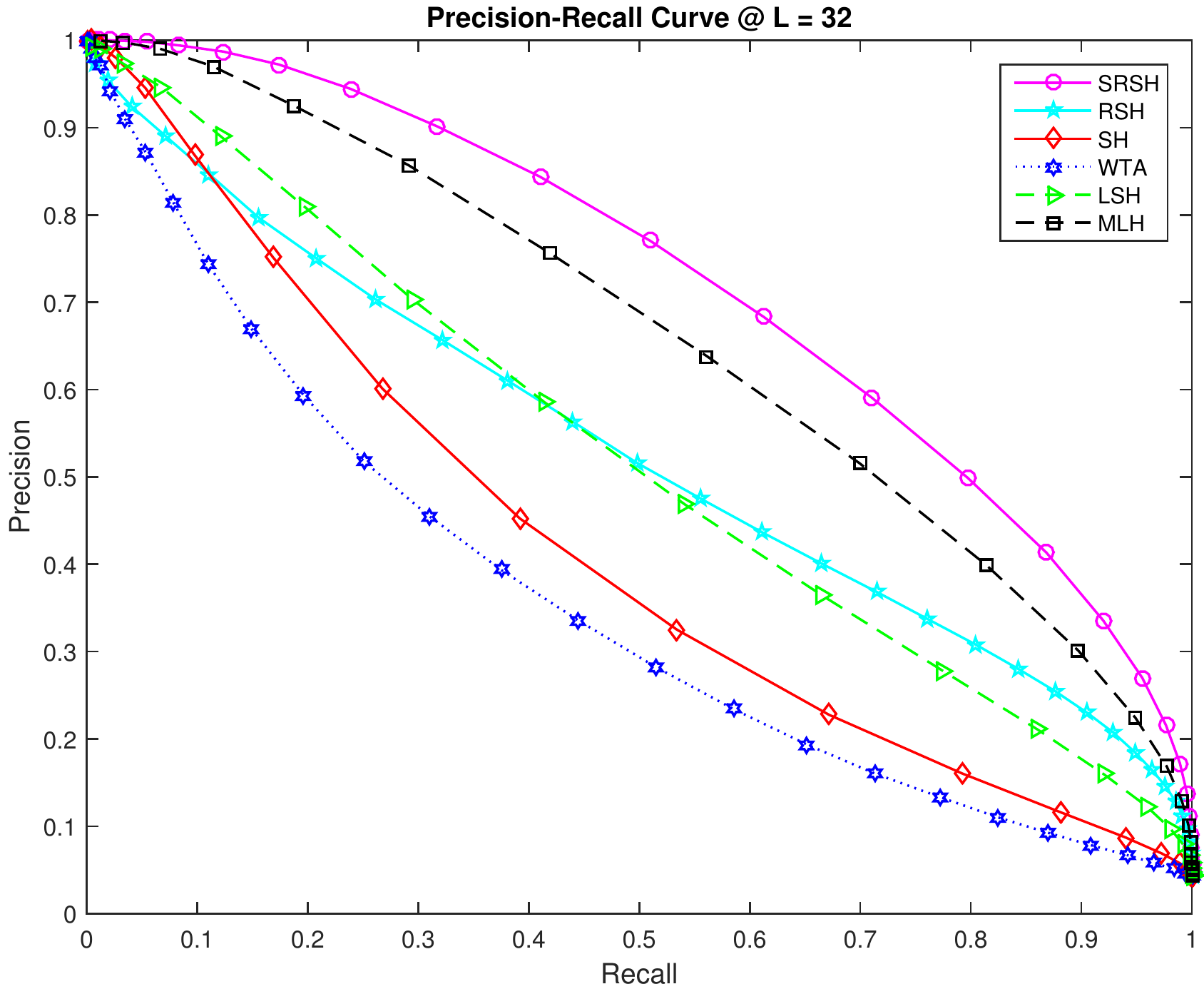}}
\subfigure[Peekaboom]{\label{prrc4}\includegraphics[width=0.3\textwidth]{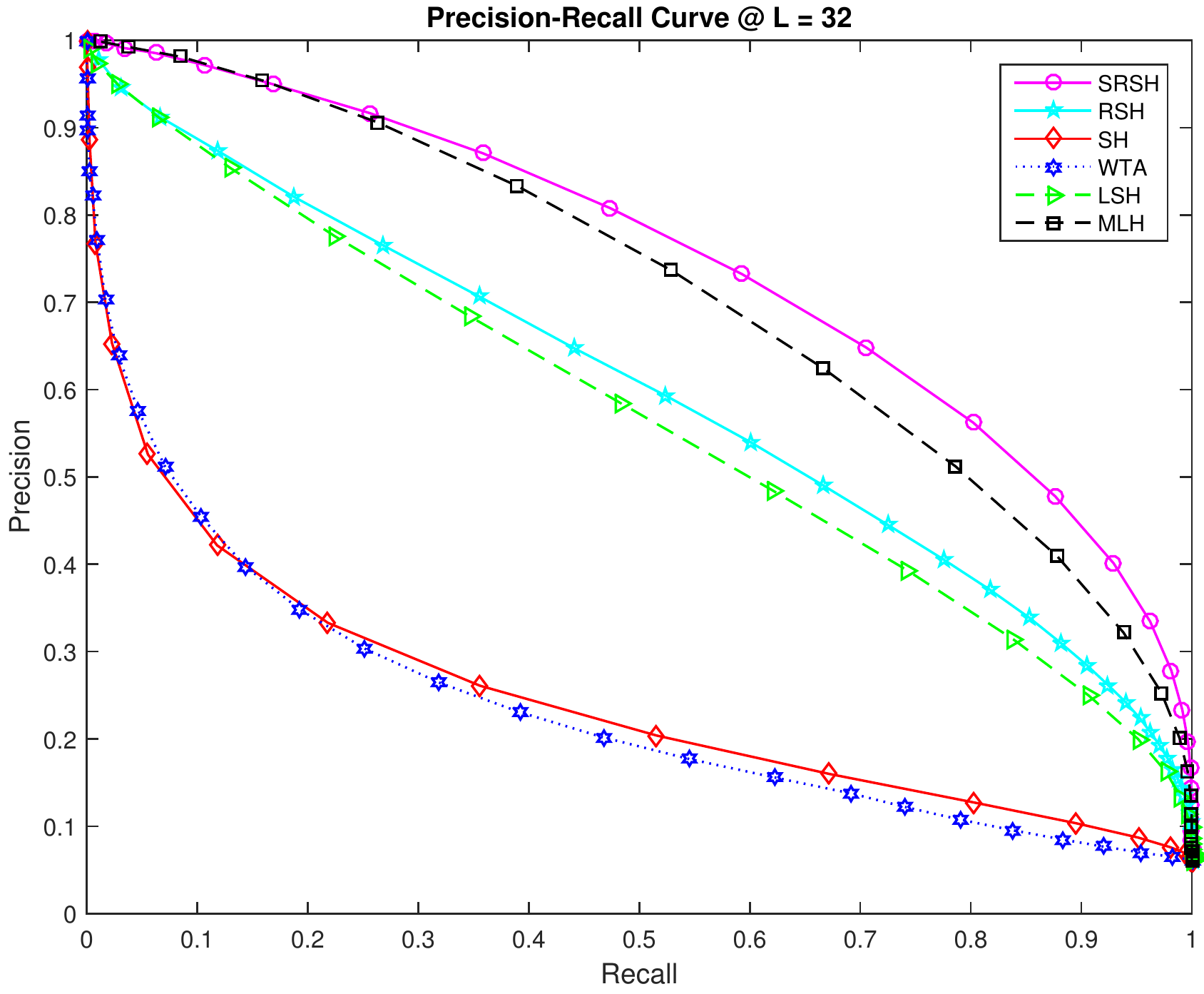}}
\caption{Precision-recall curve when code length L = 32.}
\label{prcrec}
\vskip -0.2in
\end{figure*}

\subsection{Dataset and Compared Methods}
In order to evaluate the proposed hashing approaches, Rank Subspace Hashing (RSH) and Sequential Rank Space Hashing (SRSH), we use three well-known datasets: LabelMe and Peekaboom, two collections of images represented as 512D Gist vectors designed for object recognition tasks; and MNIST, a corpus of  handwritten digits in $24\times 24$ greyscale image. The above datasets are assembled by \cite{bre} and also used in \cite{mlh}. 

Following the settings of \cite{mlh}, we  randomly picked 1000 points for training and a separate set of 3000 points as test queries. The groundtruth neighbors for test queries are defined by thresholding the Euclidean distance such that each query point has an average of 50 neighbors. Similarly, we define the neighbors and non-neighbors of each data point in the training set in order to create the similarity matrix. All the datasets are mean-centered and normalized prior to training and testing. Some methods (e.g. SH) often perform better after dimensionality reduction, we therefore apply PCA to all datasets and retain the top 40 directions for a fair comparison.

For comparison, we choose several state-of-the-art methods: Minimal Loss Hashing (MLH \cite{mlh}), Spectral Hashing (SH \cite{sh}), Locality Sensitive Hashing (LSH \cite{lsh}), and Winner-Take-All (WTA \cite{wta}) hash. For MLH and SH, we use the publicly available source code provided by their original authors, while we implemented our own version of LSH and WTA since they are rather straightforward to implement. Those methods cover both supervised (e.g., MLH) and unsupervised (e.g., SH) hashing as well as data-agnostic ones (e.g., LSH and WTA), and are considered most representative in their own category.

\begin{figure*}[t]
\vskip 0.2in
\centering
\subfigure[LabelMe]{\label{pr21}\includegraphics[width=0.3\textwidth]{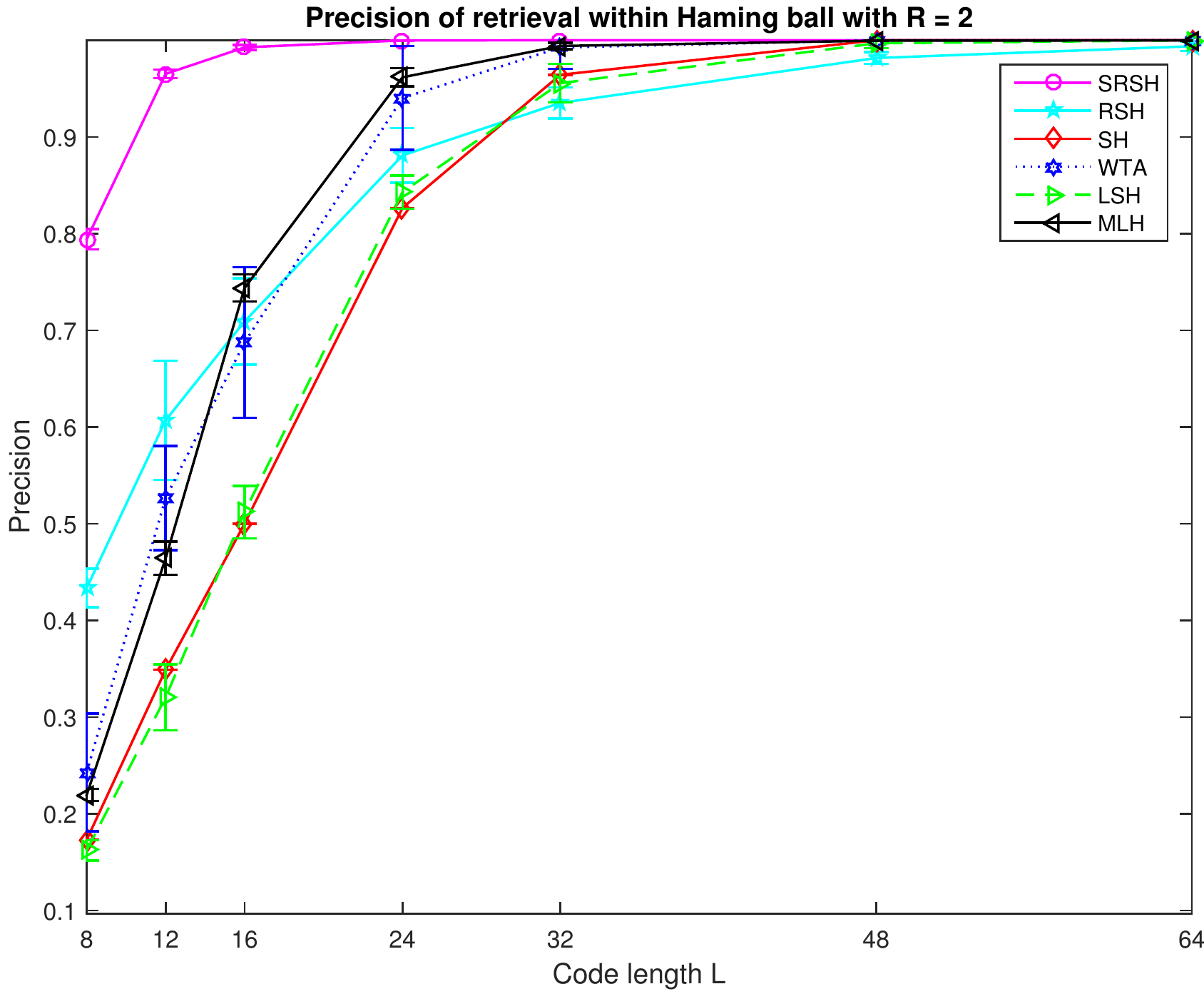}}
\subfigure[MNIST]{\label{pr22}\includegraphics[width=0.3\textwidth]{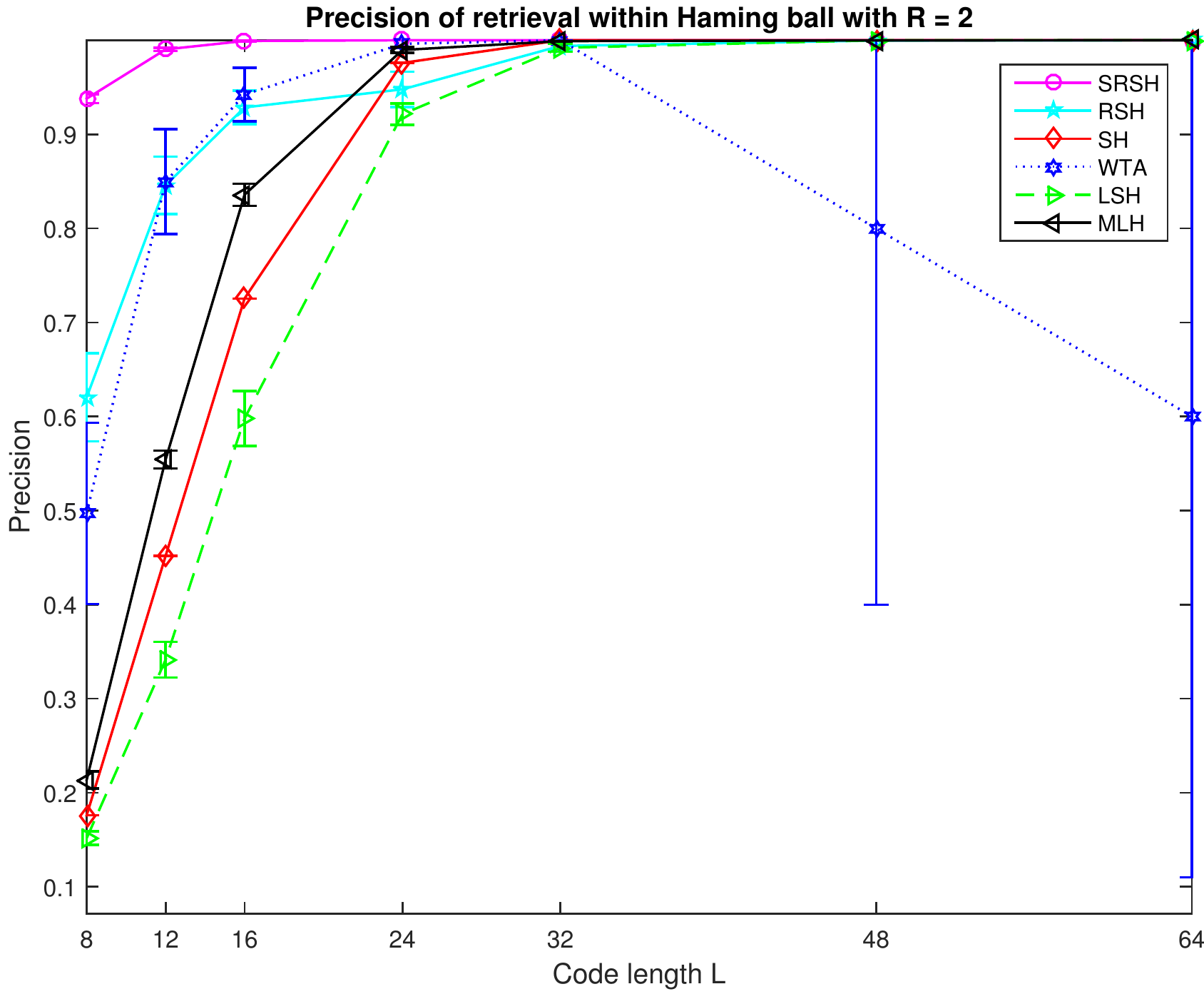}}
\subfigure[Peekaboom]{\label{p24}\includegraphics[width=0.3\textwidth]{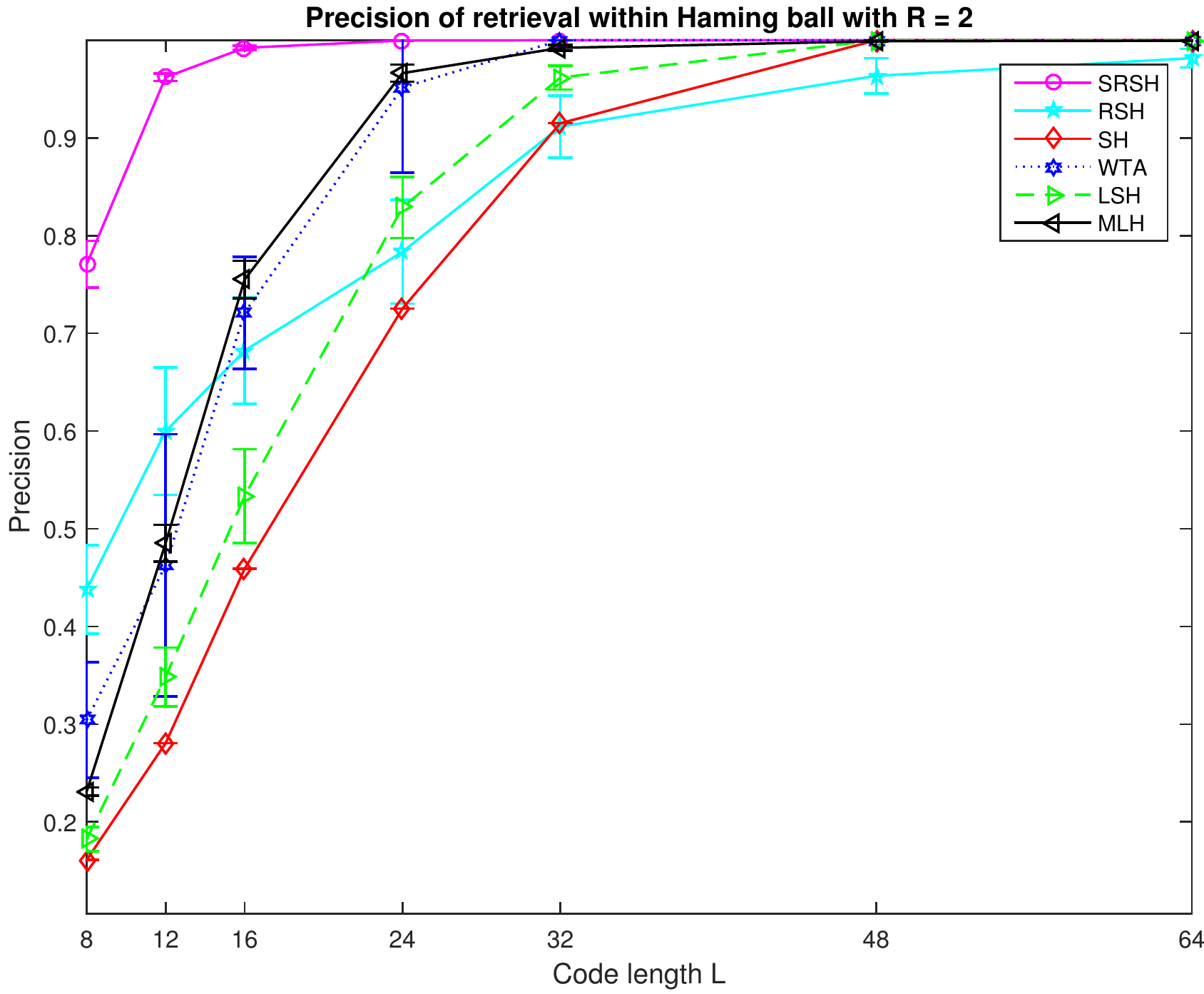}}
\caption{Precision of retrieval within a Hamming ball of radius R = 2.}
\label{pr2}
\vskip -0.2in
\end{figure*}

\begin{figure*}[t]
\vskip 0.2in
\centering
\subfigure[LabelMe]{\label{pr31}\includegraphics[width=0.3\textwidth]{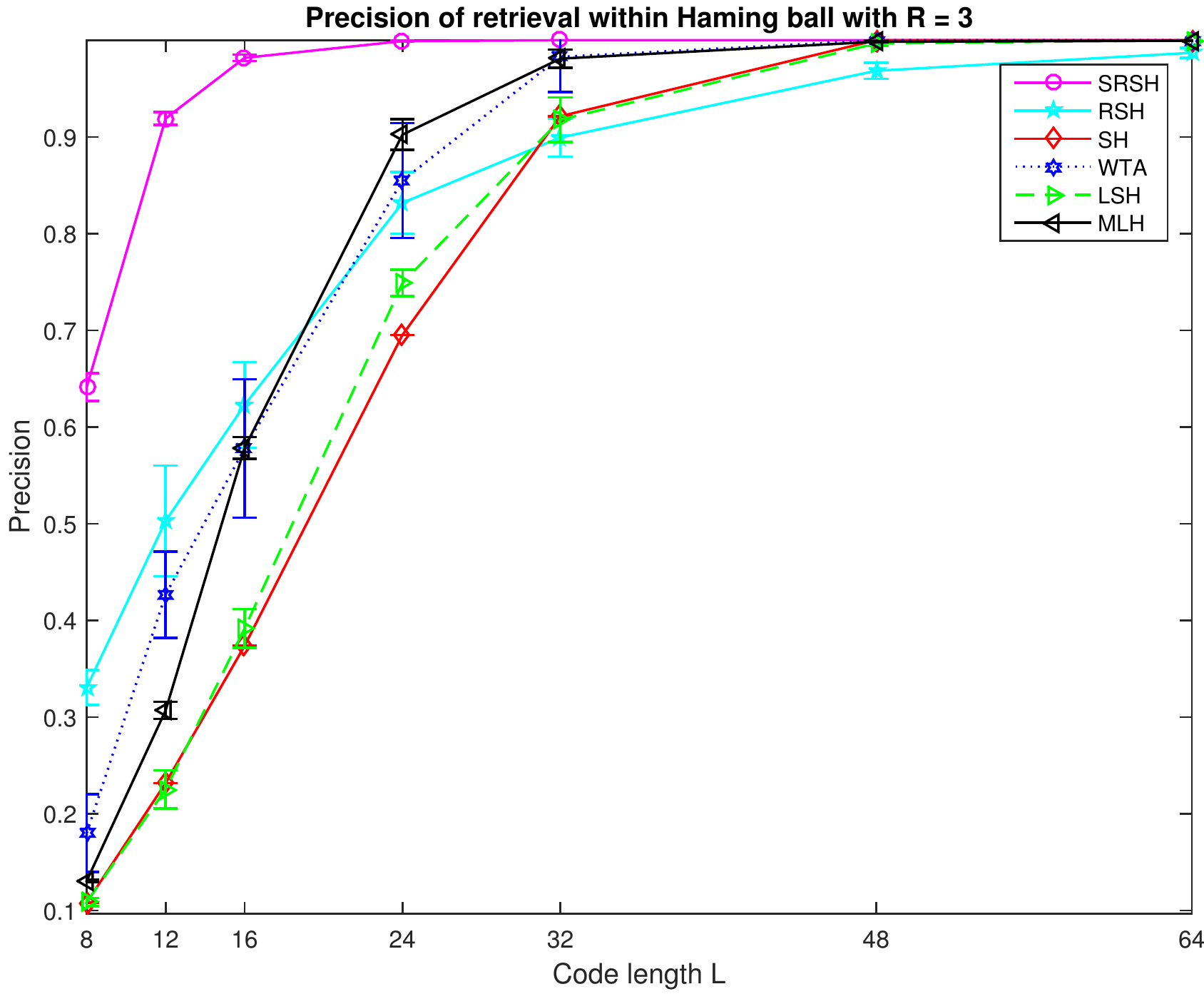}}
\subfigure[MNIST]{\label{pr32}\includegraphics[width=0.3\textwidth]{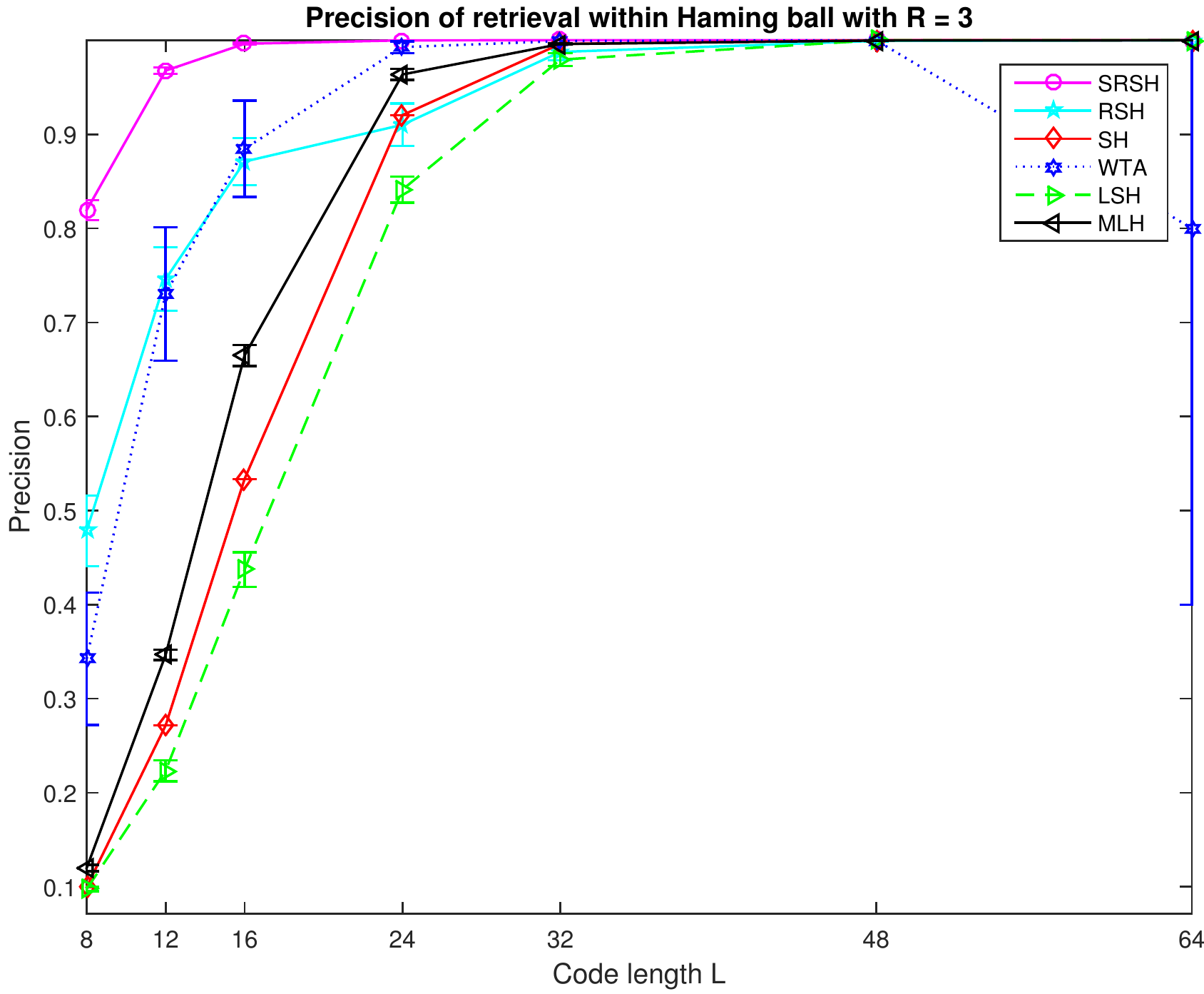}}
\subfigure[Peekaboom]{\label{p34}\includegraphics[width=0.3\textwidth]{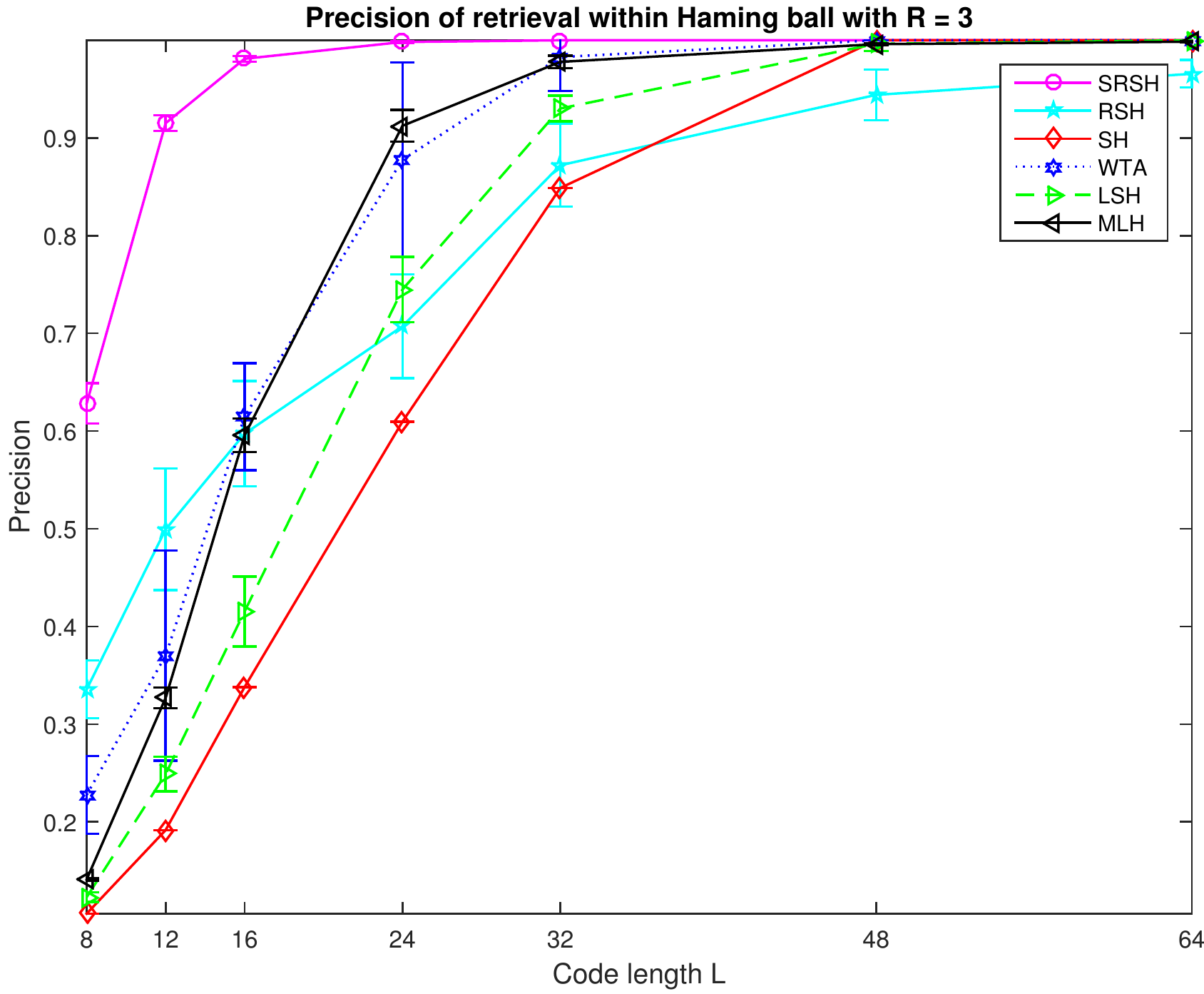}}
\caption{Precision of retrieval within a Hamming ball of radius R = 3.}
\label{pr3}
\vskip -0.2in
\end{figure*}

\subsection{Methodology}
In evaluating Approximate Nearest Neighbor (ANN) search, two methods are frequently adopted in the literature, that is, hash table based lookup and Hamming distance based kNN search. We use both methods in our evaluation. In hash table lookup, the hash code is used to index all the points in a database, and the data points with the same hash key fall into the same bucket. Typically, hash buckets that fall within a Hamming ball of radius $R$ (i.e. the hash code differs by only 2 or 3 bits) of the target query are considered to contain relevant query results. A big advantage of hash table lookup lies in that it can be done in constant time. In contrast, Hamming distance based kNN search performs a standard kNN searching procedure based on Hamming distance which involves a linear scan of the entire database. However, since Hamming distance can be computed efficiently, the kNN search in Hamming space is also very fast in practice.

In our experiments, we evaluate the retrieval quality by setting $R=2 \mbox{ and } 3$ in the hash table lookup and $k = 50 \mbox{ and } 100$ in Hamming distance based kNN search. For both evaluation protocols, we compute the retrieval precision that is defined as the percentage of true neighbors among those returned by the query. The precision reflects the quality of hash codes to a large extent and it can be critical for many applications. In addition, we also evaluate the average precision for hash table lookup by varying $R$, which approximates the area under precision-recall curve. For all benchmarks (unless otherwise specified), we run every algorithm 10 independent times and report the mean and the standard deviation.

As for parameter settings, MLH requires a loss scaling factor $\epsilon$, and two loss function hyper-parameters $\rho$ and $\lambda$. We follow the practice of MLH and perform cross-validation on a number of combinations to get the best performance at each code length. Similarly, our algorithm also has three hyper-parameter, that is, the subspace dimension $K$ and two error term hyper-parameters as defined in (\ref{err}). A similar cross validation procedure is used to find the best model. For WTA, we use the polynomial kernel extension and set window size $K = 4$ and polynomial degree $p = 4$, as suggested by \cite{wta}. SH and LSH are essentially parameter free and therefore do not require special handling.

\subsection{Results}
Figure \ref{ap} shows the average precision using different hash code length. We aim to compare the performance of different hashing methods in generating compact hash code, therefore the code length is restricted below 64. It can be observed from the Figure \ref{ap} that the average precision of almost all methods increases monotonically when codes become longer, which is reasonable since longer codes retains more information of original data. The only exception to this trend is SH whose performance doesn't increase or even slightly drops after exceeding certain number of bits (e.g. 24 to 32). This can be explained by fact that unsupervised learning methods tend to overfit more easily with longer codes, which is consistent with the observation by \cite{splh}. 

We note that RSH shows significant improvement over WTA, another representative ranking-based hashing algorithm, as a result of the generalization of projection directions and the supervised learning process. Compared with RSH, SRSH further boosts the performance with large gains across all the tested datasets, demonstrating the effectiveness of the sequential learning method. In general, SRSH achieves the best performance, with about $10\%$ lead over MLH. We also note that both of our algorithms demonstrate exceptional performance with extremely short code (e.g. of length less than 12) as a result of using rank order encoding.

\begin{figure*}[ht]
\vskip 0.2in
\centering
\subfigure[LabelMe]{\label{pm501}\includegraphics[width=0.3\textwidth]{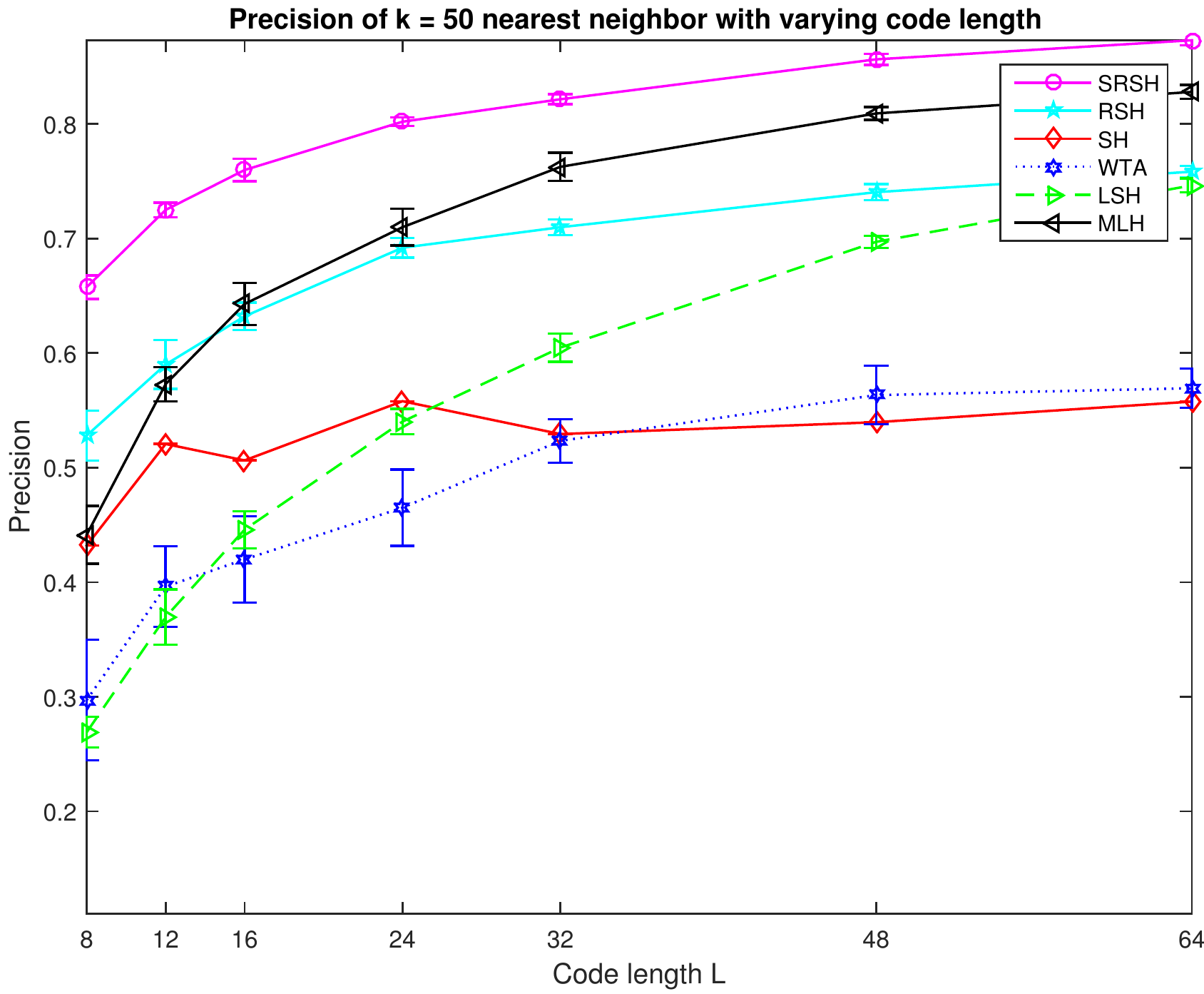}}
\subfigure[MNIST]{\label{pm502}\includegraphics[width=0.3\textwidth]{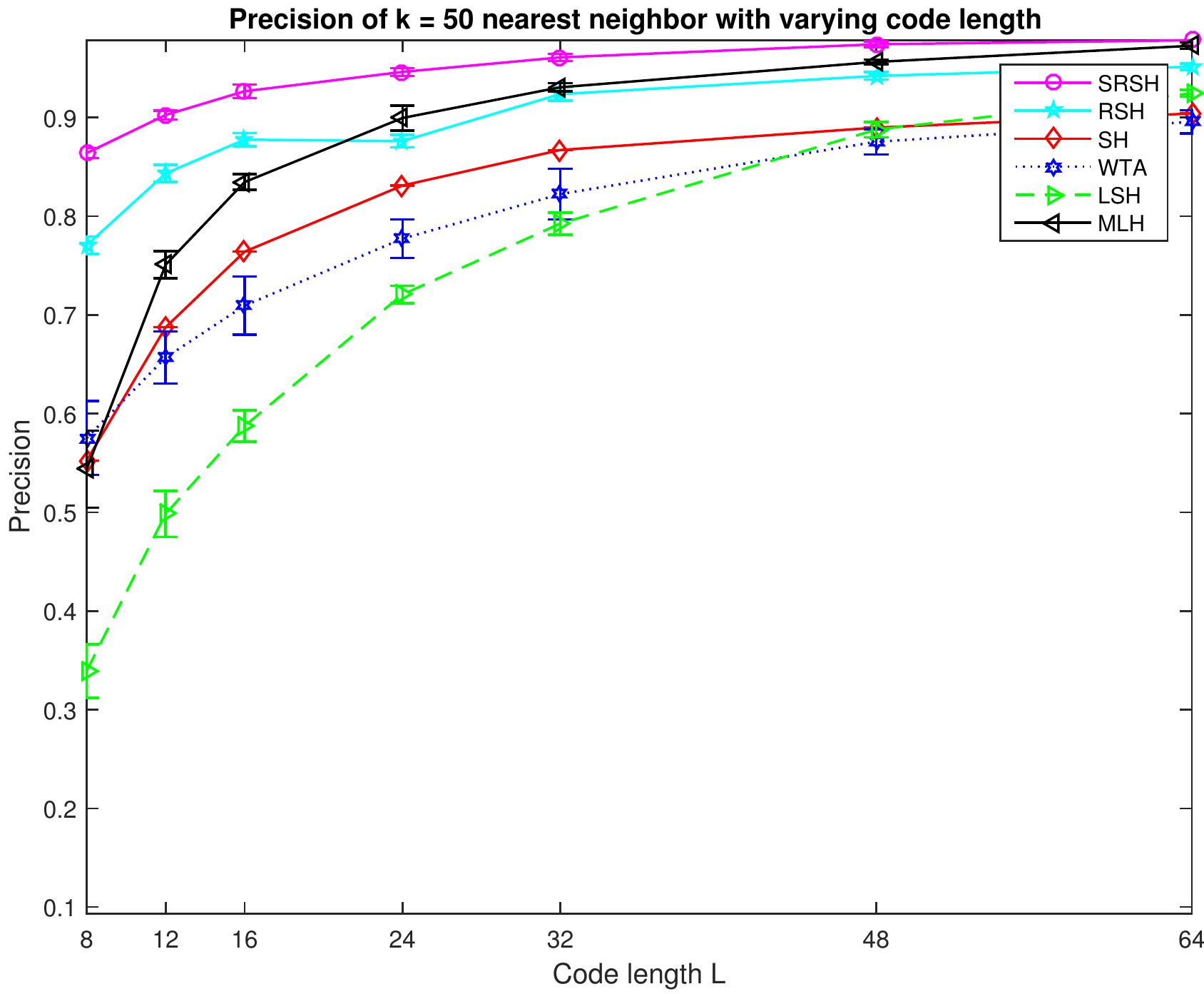}}
\subfigure[Peekaboom]{\label{pm504}\includegraphics[width=0.3\textwidth]{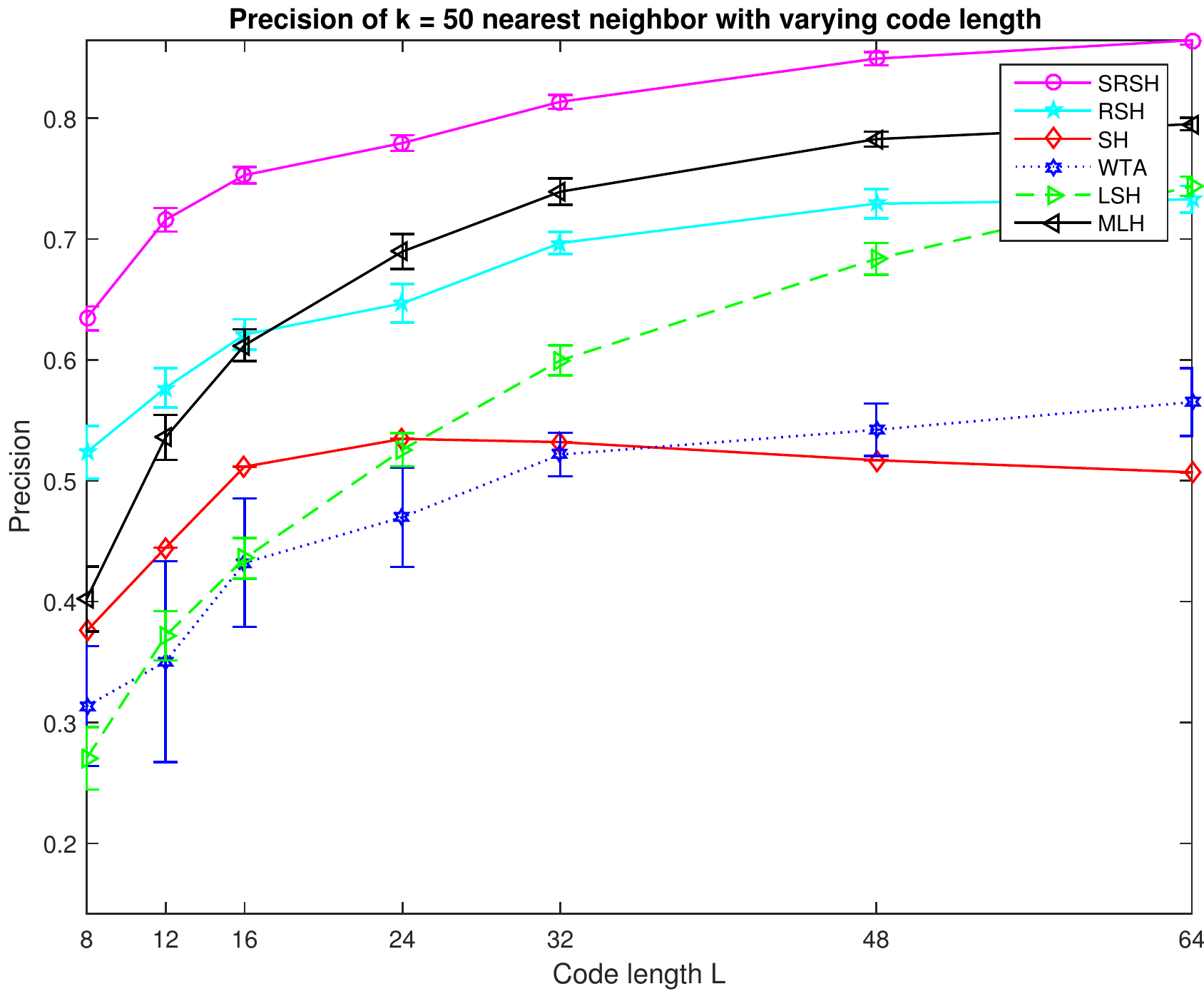}}
\caption{Precision of 50 nearest neighbor retrieval based on Hamming distance.}
\label{pm50}
\vskip -0.2in
\end{figure*}

\begin{figure*}[ht]
\vskip 0.2in
\centering
\subfigure[LabelMe]{\label{pm1001}\includegraphics[width=0.3\textwidth]{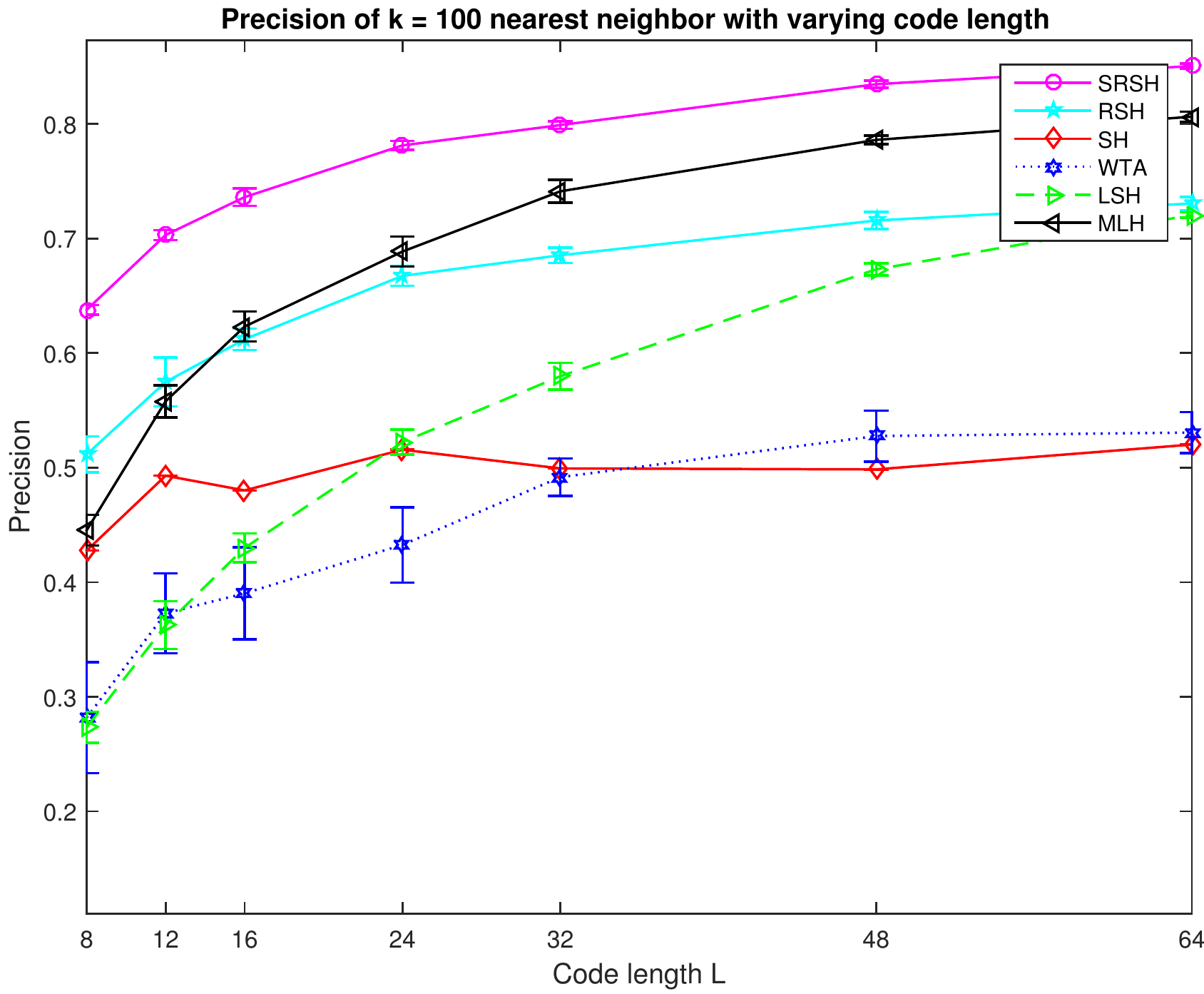}}
\subfigure[MNIST]{\label{pm1002}\includegraphics[width=0.3\textwidth]{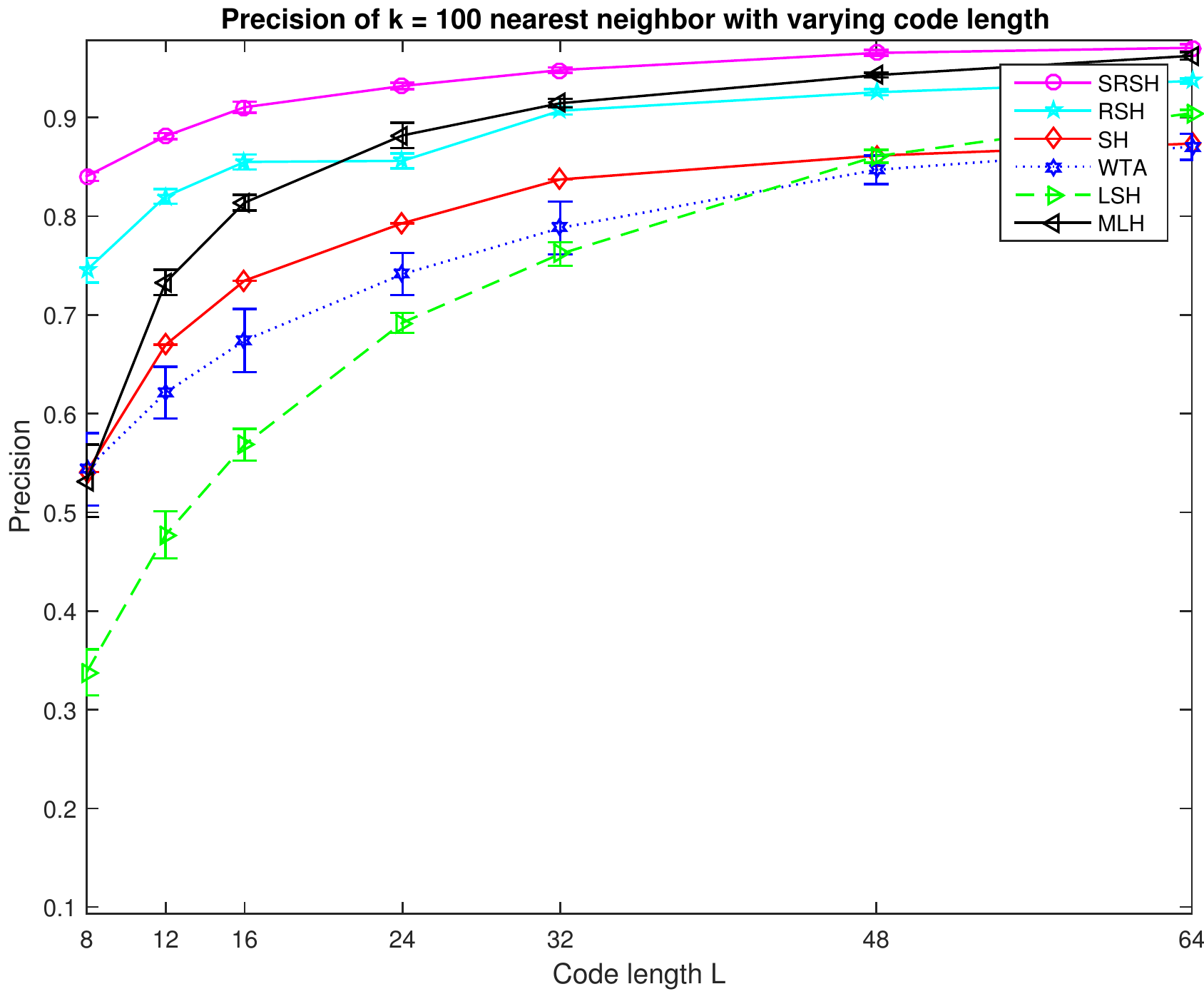}}
\subfigure[Peekaboom]{\label{pm1004}\includegraphics[width=0.3\textwidth]{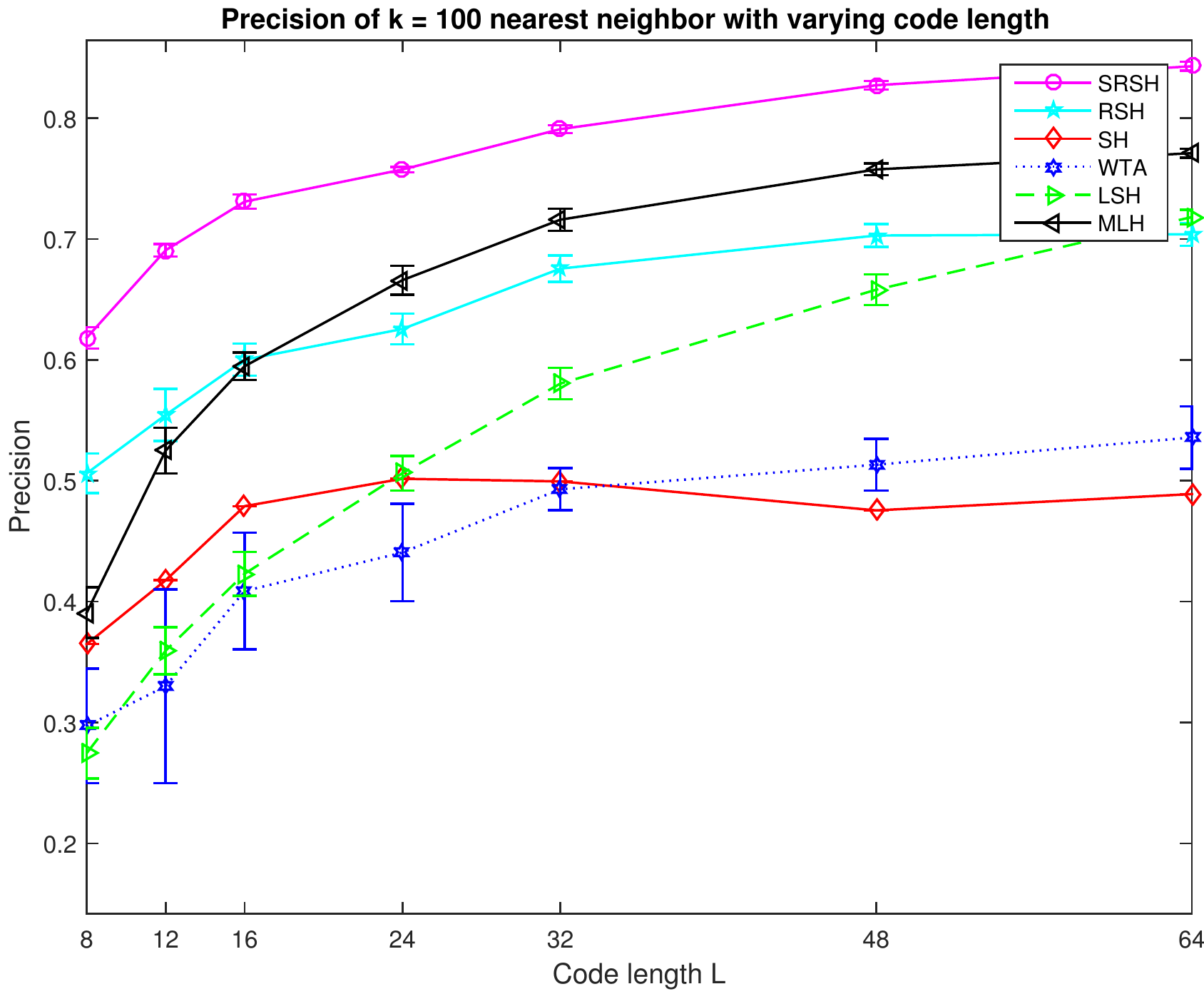}}
\caption{Precision of 100 nearest neighbor retrieval based on Hamming distance.}
\label{pm100}
\vskip -0.2in
\end{figure*}

In addition to the average precision, we also show a more detailed precision-recall profile when the code length $L$ is fixed to 32 in Figure \ref{prcrec}. In the precision-recall curve, better performance is shown by larger area under the curve. Again, both of our algorithms perform significantly better than WTA with SRSH consistently being the best, which is consistent with the previous results.

The results of hash table lookups are shown in Fig. \ref{pr2} and Fig. \ref{pr3}, for $R = 2$ and $R = 3$ respectively. As explained in the previous section, precision alone is more critical than average precision that is an overall evaluation of both precision and recall. Therefore, the results in Figure \ref{pr2} and Figure \ref{pr3} can be more important for such applications. In those tests, rank order based techniques (i.e. WTA, RSH and SRSH) generally perform better than numeric value based hashing schemes because of certain degree of resilience to numeric noises/perturbations. For example, although both WTA and LSH are based on data-agnostic random methods, WTA clearly outperforms LSH for most of the tests, which is similar to the results obtained in (\cite{wta}). However, we find that WTA sometimes fails to retrieve any neighbor within a small Hamming ball, resulting in large standard deviation in precision at large code length (e.g. Fig. \ref{pr22} and \ref{pr32}). This is a natural result of applying randomness to a highly selective hash function. Such limitation is effectively addressed by providing certain supervision in obtaining the hash functions. Therefore, both RSH and SRSH produce more stable results than WTA, as demonstrated by the consistently smaller standard deviations. Overall, SRSH performs the best in all the tests, again demonstrating its superiority in generating high quality hash codes.

The last group of experiments is the Hamming distance based kNN search, where we evaluate the precision of true neighbors among the 50 and 100 nearest neighbors measured by Hamming distance. As shown in Figure \ref{pm50} and Figure \ref{pm100}, the  results are similar to those in the hash table lookups, except that there are no missed retrievals for any of the compared algorithms because all queries are guaranteed to return the specified number of results. The proposed algorithms both give competitive results as compared with the others.

\section{Conclusion}
\label{conclude}
In this paper, a new reformulation of the Winner-Take-All hashing scheme is first presented. Based on this formulation, we propose a novel hash learning objective that aims to optimize a number of low-dimensional linear subspaces for high quality rank order-based hash encoding. A simple yet effective learning algorithm is then provided to optimize the objective function, leading to a number of optimal rank subspaces. The effectiveness of the proposed learning method in addressing the limitations of WTA is verified in a number of experiments. We also embed our learning method into a sequential learning framework that pushes the performance of the basic learning algorithm even further. Extensive experiments on several well-known datasets demonstrated our superior performance over state-of-the-art.

\nocite{langley00}

\bibliography{reference}
\bibliographystyle{icml2015}

\end{document}